\documentclass{article}
\usepackage[final]{neurips_2020}
\usepackage[utf8]{inputenc}
\usepackage[T1]{fontenc}
\usepackage{hyperref}
\usepackage{url}
\usepackage{booktabs}
\usepackage{amsfonts}
\usepackage{nicefrac}
\usepackage{microtype}
\usepackage{booktabs}
\usepackage[ruled,linesnumbered,noend]{algorithm2e}
\usepackage{comment}
\usepackage{amsmath}
\usepackage{cleveref}
\usepackage{longtable}
\usepackage{pdflscape}
\usepackage{mathtools}
\usepackage{array}
\usepackage{color}
\usepackage{subcaption}
\usepackage[symbol]{footmisc}

\usepackage{varwidth}

\DeclareMathOperator*{\argmax}{arg\,max}
\DeclareMathOperator*{\argmin}{arg\,min}
\DeclarePairedDelimiterX{\infdivx}[2]{(}{)}{%
  #1\;\delimsize\|\;#2%
}
\newcommand{\infdiv}{KL\infdivx}
\newcommand{\cmt}[1]{\ignorespaces}
\newcolumntype{H}{>{\setbox0=\hbox\bgroup}c<{\egroup}@{}}

\newcommand{\methodname}{Automatic Recall Machines\xspace}
\newcommand{\methodnameshort}{ARM\xspace}

\newcommand\extrafootertext[1]{%
    \bgroup
    \renewcommand\thefootnote{\fnsymbol{footnote}}%
    \renewcommand\thempfootnote{\fnsymbol{mpfootnote}}%
    \footnotetext[0]{#1}%
    \egroup
}

\title{\methodname: \\Internal Replay, Continual Learning and the Brain}

\author{
    {\bfseries Xu Ji$^{1}$, Jo\~ao Henriques$^{1}$, Tinne Tuytelaars$^{2}$, Andrea Vedaldi$^{1}$}\vspace{0.2cm} \\
    $^1$VGG, University of Oxford \\ 
    $^2$KU Leuven \\
}

\begin{document}
\maketitle

\begin{abstract}
Replay in neural networks involves training on sequential data with memorized samples, which counteracts forgetting of previous behavior caused by non-stationarity.
We present a method where these auxiliary samples are generated on the fly, given only the model that is being trained for the assessed objective,
without extraneous buffers or generator networks. Instead the implicit memory of learned samples within the assessed model itself is exploited. 
Furthermore, whereas existing work focuses on reinforcing the full seen data distribution, we show that optimizing for not forgetting calls for the generation of samples that are specialized to each real training batch, which is more efficient and scalable.
We consider high-level parallels with the brain, notably the use of a single model for inference and recall, the dependency of recalled samples on the current environment batch, top-down modulation of activations and learning, abstract recall, and the dependency between the degree to which a task is learned and the degree to which it is recalled.
These characteristics emerge naturally from the method without being controlled for.
\end{abstract}

\section{Introduction}\label{s:intro}
The ability to learn effectively from sequential or non-stationary data remains a fundamental difference between human brains and artificial neural networks. 
The development of biological brains is impaired when the environment changes too fast~\citep{wood2016smoothness}.
On the other hand, it has long been observed that artificial neural networks catastrophically forget previously learned behavior when data is presented in a sequential manner~\citep{mccloskey1989catastrophic}.
In other words, while biological learning favors datastreams that focus on one task at a time, the effectiveness of stochastic gradient descent depends on the opposite quality of exposure to all tasks within a small temporal window. 


The retention of learned behavior in neural networks can be increased with experience replay, enforced non-distributedness, or parameter-level regularization.
Of these, replay is perhaps the most promising approach.
Parameter-level regularization has been shown to underperform~\citep{chaudhry2019continual, van2019three}, likely due to restricting change with excessive stringency~\citep{ramapuram2020lifelong}.
Non-distributed representations limit parameter sharing a priori, preventing interference but sacrificing on memory efficiency and transfer~\citep{french1991using, bengio2013representation}.

Replay, where training is augmented with auxiliary batches intended to capture previously seen data~\citep{robins1995catastrophic}, avoids both these issues as it allows for constraints to be imposed at functional level over a distributed representation.
However, existing work generally uses buffers or generator networks to memorize the seen data~\citep{chaudhry2019continual, shin2017continual}. Buffers are memory inefficient, and generator networks have proved difficult to train for natural images given only sequential data~\citep{lesort2019generative,aljundi2019online}.
In both cases it is also inconvenient to require additional memory for data memorization, especially as the model being trained for directly pertinent tasks already has an implicit memory of past training samples.

\begin{figure}
\centering
\includegraphics[width=0.99\linewidth]{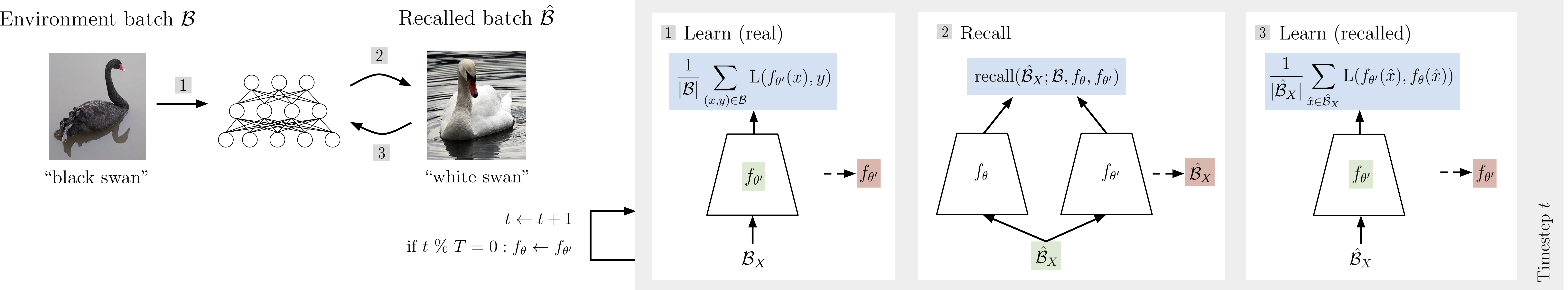}
\setlength{\belowcaptionskip}{-1em}
\caption{\methodname. 
In each stage, the model or recalled samples (green) are optimized for objective (blue) and updated (red).
Reinforcing the resulting maximally interfered
$\mathcal{\hat{B}}_X$ with old outputs from $f_{\theta}$ optimally minimizes forgetting of $f_{\theta'}$,
because in theory if the maximal divergence between old and new non-local behavior is 0, then all such divergences are 0. 
Thus training on $\mathcal{B}$ updates local behavior while training on $\mathcal{\hat{B}}$ protects behavior in the rest of input space.
}
\label{f:splash}
\end{figure}

The goal of this work, \methodname (\methodnameshort), is to optimally exploit the implicit memory in the tasks model for not forgetting, by using its parameters for both inference and generation~(\cref{f:splash}).
Instead of attempting to reproduce the full seen data distribution, we generate specifically the optimal samples conditioned on the current real batch, which is more efficient and scalable.
Our derivation shows that these samples are the points
in input space whose outputs are maximally changed given training on the real batch, providing an explanation for why training with the most dissonant related sample is optimal for not forgetting, an intuition that was used in~\cite{aljundi2019online}. Memory reconsolidation in humans also appears to favor conflicting experiences~\citep{sinclair2018surprise}.

We use \emph{recall} to refer to internally generated replay. \emph{Automatic} refers to recall being a direct consequence of learning on the environment.
Without an initial learning step, there is no change in knowledge held by the network to compensate for and thus no recall; vice-versa, environmental batches that cause a significant change in network knowledge induce more learning from recall~(\cref{e:best_params}). 
In the brain, experiences associated with higher surprise and thus representational change also increase the intensity of replay during and after the experience~\citep{cheng2008new,o2008reactivation}.
\begin{figure}[b]
\centering
\includegraphics[width=0.8\linewidth]{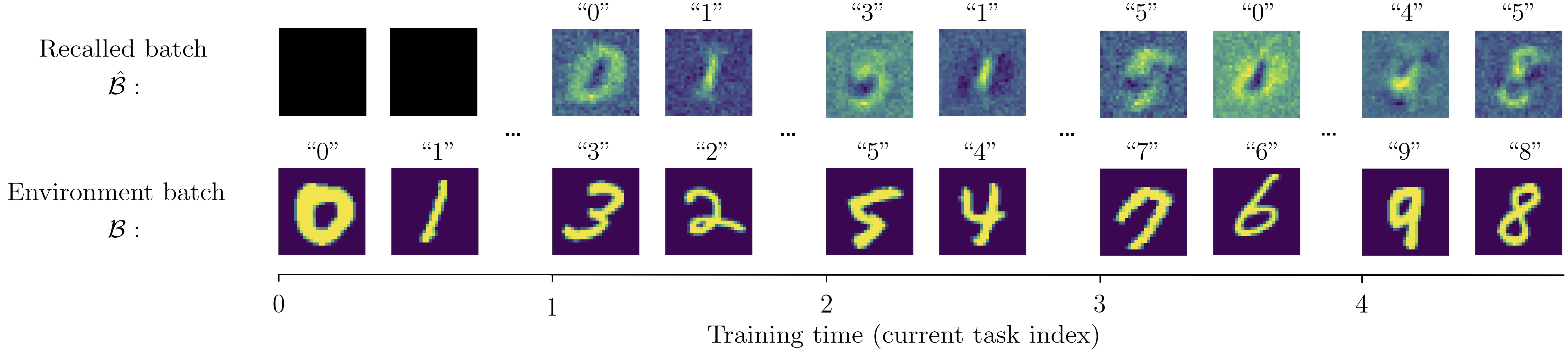}
\setlength{\belowcaptionskip}{-0.75em}
\caption{
Real $\mathcal{B}$ and generated $\mathcal{\hat{B}}$ for MNIST. Images in the top row were initialized from the image below before being optimized by $\operatorname{recall}$~(\cref{e:recall}).
}
\label{f:mnist_prog}
\end{figure}
\section{\methodname}

Given the space of outputs, $\mathcal{Y} = \{1,\dots,C\}$, and space of inputs, for example images $\mathcal{X} = [0,1]^{3 \times H \times W}$, define the knowledge of the categorical neural network $f_{\theta}$ as its graph:
\begin{align}\label{e:k}
\operatorname{graph}(f_\theta)=
\{ (x,y) \in \mathcal{X} \times \mathcal{Y} : f_\theta(x) = y \}.
\end{align}
When training on real batch $\mathcal{B} \subset \mathcal{D}$ from non-stationary datastream $\mathcal{D} \subset \mathcal{X}\times\mathcal{Y}$ to obtain updated parameters $\theta'$, we would like the difference between old $\theta$ and new $\theta'$ to amount to a \emph{local} change in knowledge. 
$\theta$ could be the snapshot of parameters immediately preceding $\theta'$ or, more generally, preceding $\theta'$ by $T \geq 1$ timesteps, since we make no assumption that $T = 1$.
Let $\mathcal{B}_X$ and $\mathcal{B}_Y$ denote the inputs and output targets of $\mathcal{B}$.
Define knowledge preservation or not forgetting as:
\begin{align}\label{e:kp_orig}
\forall \hat{x}\in\mathcal{X} ~.~
\neg \operatorname{local}(\hat{x}; \mathcal{B}, f_{\theta}, f_{\theta'}):
f_{\theta}(\hat{x}) = f_{\theta'}(\hat{x}&) \\ 
\text{where } \operatorname{local}(\hat{x}; \mathcal{B}, f_{\theta}, f_{\theta'}) = f_{\theta}(\hat{x}) \in \mathcal{B}_Y ~\vee~ f_{\theta'}(\hat{x}) & \in \mathcal{B}_Y. \label{e:local_def}
\end{align}
This states that predictions for the input space should be the same for $f_{\theta}$ and $f_{\theta'}$, except inputs mapped to the classes in $\mathcal{B}$ i.e. classes currently being trained. \Cref{e:kp_orig} is satisfied iff:
%
\begin{align}\label{e:max_zero}
\max_{\substack{\hat{x}\in\mathcal{X} \\ \neg \operatorname{local}(\hat{x}; \mathcal{B}, f_{\theta}, f_{\theta'})}} \operatorname{D}(f_{\theta'}(\hat{x}), f_{\theta}(\hat{x})) = 0,
\end{align}
where $\operatorname{D}$ indicates divergence; we use the symmetric Jensen-Shannon divergence. 
From \cref{e:max_zero}, we see that in order to suppress the forgetting of non-local knowledge, the maximally violating input should result in no violation. Hence the best knowledge preserving parameters in each training iteration on original batch $\mathcal{B}$ can be given by:
\begin{align}
\theta^* = \argmin_{\theta'} 
\Bigg[
\Bigg(
\frac{1}{N} \sum_{(x, y) \in \mathcal{B}} \operatorname{L}(f_{\theta'}(x),y)
\Bigg)
~+
\sup_{\substack{\hat{x}\in\mathcal{X} \\ \neg \operatorname{local}(\hat{x}; \mathcal{B}, f_{\theta}, f_{\theta'})}} \operatorname{D}(f_{\theta'}(\hat{x}), f_{\theta}(\hat{x}))
\Bigg],
\end{align}
where $\operatorname{L}$ denotes the original tasks loss, which is cross-entropy in our experiments. 
This can be approximated by:
\begin{align}\label{e:best_params}
\theta^* = \argmin_{\theta'} 
\Bigg[
\Bigg(
\frac{1}{N} &\sum_{(x, y) \in \mathcal{B}} \operatorname{L}(f_{\theta'}(x),y)
\Bigg)
~+~
\lambda_0 \operatorname{L}(f_{\theta'}(\hat{x}^*), f_{\theta}(\hat{x}^*))
\Bigg] \\
\text{where } 
\hat{x}^* &= \argmax_{\substack{\hat{x}\in\mathcal{X} \\ \neg \operatorname{local}(\hat{x}; \mathcal{B}, f_{\theta}, f_{\theta'})}} \operatorname{D}(f_{\theta'}(\hat{x}), f_{\theta}(\hat{x})),\label{e:max_int_input}
\end{align}
with $\hat{x}^*$ being the maximally violating input. 
The key idea is replaying maximally violating inputs with their old targets minimizes the maximal divergence between old model $f_\theta$ and new model $f_{\theta'}$, thus optimizing for not forgetting via \cref{e:max_zero}.
Note no gradient exists for $\hat{x}^*$ until $\theta' \neq \theta$. 
This is what is meant by \emph{automatic} (\cref{s:intro}); the optimization signal during recall~(\cref{e:max_int_input}) and from recall~(\cref{e:best_params}) scales with the change between $\theta$ and $\theta'$. 
In practice, we fill a full replay batch of $M > 1$ images, $\mathcal{\hat{B}}_X$, by maximising:
\begin{small}
\begin{align}
&\operatorname{recall}(\mathcal{\hat{B}}_X;\mathcal{B}, f_\theta, f_{\theta'}) = \nonumber
\\
&
\Bigg[
\frac{1}{M} \sum_{\hat{x} \in \mathcal{\hat{B}}_X} 
\Bigg(
\operatorname{D}(f_{\theta'}(\hat{x}),f_{\theta}(\hat{x}))
+
\frac{\lambda_1}{C} \sum_{y \in \operatorname{set}(\mathcal{B}_Y)} \operatorname{L}(f_{\theta}(\hat{x}), y) 
+
\frac{\lambda_2}{C} \sum_{y \in \operatorname{set}(\mathcal{B}_Y)} \operatorname{L}(f_{\theta'}(\hat{x}), y) 
\Bigg)
\Bigg]
+ \operatorname{reg}(\mathcal{\hat{B}}_X; f_{\theta}). \label{e:recall}
\end{align}
\end{small}
\hspace{-0.58em} This is 
\cref{e:max_int_input} applied to each element of the batch with the constraint on $\operatorname{local}$ converted into regularization terms.
$\mathcal{\hat{B}}_Y = [f_{\theta}(\hat{x}) : \hat{x} \in \mathcal{\hat{B}}_X ]$, $\operatorname{set}(\mathcal{B}_Y) = \{y : y \in \mathcal{B}_Y\}$ and $C=|\operatorname{set}(\mathcal{B}_Y)|$.
Regularization across the batch is provided by $\operatorname{reg}$:
\begin{small}
\begin{align}
\operatorname{reg}(\mathcal{\hat{B}}_X; f_{\theta}) =
\lambda_3 \mathcal{H}(\mathcal{\hat{B}}_Y) 
-
\Bigg(
\frac{\lambda_4}{M} \sum_{\hat{y} \in \mathcal{\hat{B}}_Y}  \operatorname{L}(\hat{y}, \operatorname{argmax}(\hat{y}))
\Bigg)
-
\lambda_5 L2(\mathcal{\hat{B}}_X) - 
\lambda_6 TV(\mathcal{\hat{B}}_X),\label{e:reg}
\end{align}
\end{small}
\hspace{-0.3em}comprising of entropy maximisation with sharpening to prevent duplication, and L2 norm and total variation minimization to smooth inputs~\citep{ mahendran2015understanding,yin2019dreaming}.
$\mathcal{H}$ denotes entropy. 
The full training procedure is given in~\cref{algo} in Appendix.
\section{Experiments}\label{s:experiments}

\subsection{Quantitative results}\label{s:exp_quantiative}
We use standard ``class-incremental'' training and evaluation protocol on sequential CIFAR10-5, MiniImageNet-20 and MNIST-5k-5~\citep{aljundi2019online}, where suffix indicates number of tasks.
On CIFAR10 and MiniImageNet, replaying 100 recalled images with \methodnameshort outperforms experience replay (ER) with 100 real stored images, regardless of whether MIR is used for the latter~(\cref{tab:cifar10,tab:miniimagenet}).
On MiniImageNet, replaying 100 recalled images with \methodnameshort achieves comparable performance to ER with 500 stored real images, whilst more than halving the use of additional memory.

Unit lag is perhaps the most interesting variant of \methodnameshort from the perspective of biological plausibility, as the old network is the current network at the start of each training iteration, hence additional memory cost is 0.
Performance drops, as using an extremely proximal version of parameters for distillation allows for more drift.
However, \methodnameshort still outperforms ADI and standard distillation~(\cref{tab:cifar10_2}).

\begin{table}[ht]
\fontsize{6}{6}\selectfont
\setlength{\tabcolsep}{2pt}
\begin{minipage}[t]{.48\textwidth}
\vspace{0pt}
\begin{tabular}{l H c c c c c c c}
\toprule
Method & Models & \multicolumn{2}{c}{+Sample mem.} & \multicolumn{2}{c}{+Model mem.} & $M$ & Accuracy & Forgetting \\
& & \#images & MB & \#params & MB & & & \\
\midrule
Naive SGD$\dagger$ & 6462 & 0 & 0 & 0 & 0 & 0 & 55.2 $\pm$ 5.0 & - \\

\midrule

ER & - & 200 & 0.61 & 0 & 0 & 10 & 27.5 $\pm$ 1.2 &  50.5 $\pm$ 2.4 \\

ER-MIR & - & 200 & 0.61 & 0 & 0 & 10 & 29.8 $\pm$ 1.1 & 50.2 $\pm$ 2.0 \\

iCarl (5 iter) & - & 200 & 0.61 & 0 & 0 & - & 28.6 $\pm$ 1.2 & 49.0 $\pm$ 2.4 \\

GEM & - & 200 & 0.61 & 0 & 0 & - & 16.8 $\pm$ 1.1 & 73.5 $\pm$ 1.7 \\

ER & - & 100 & 0.31 & 0 & 0 & 10 & 22.4 $\pm$ 1.1 & 66.2 $\pm$ 6.2 \\

ER-MIR & - &  100 & 0.31 & 0 & 0 & 10 & 23.6 $\pm$ 0.9 & 61.4 $\pm$ 1.8 \\

\midrule

Naive SGD & 2522 & 0 & 0 & 0 & 0 & 0 & 15.0 $\pm$ 3.1 & 69.4 $\pm$ 4.7 \\

GEN & (81) & 0 & 0 & 8.63M & 34.5 & 100 & 15.3 $\pm$ 0.5 & 61.3 $\pm$ 5.1 \\

GEN-MIR & (0) & 0 & 0 & 9.50M & 38.0 & 40 & 15.3 $\pm$ 1.2 & 61.0  $\pm$ 1.2 \\

Distill (LwF) & 4082 & 0 & 0 & 1.09M & 4.38 & 100 & 19.2 $\pm$ 0.3 & 60.9 $\pm$ 3.9  \\

ADI & 2262 & 0 & 0 & 1.09M & 4.38 & 100 & 24.8 $\pm$ 0.9 & 12.0 $\pm$ 4.5 \\

\textbf{\methodnameshort} & 3717 & \textbf{0} & \textbf{0} & \textbf{1.09M} & \textbf{4.38} & \textbf{100} & \textbf{26.4 $\pm$ 1.2} & \textbf{8.07 $\pm$ 5.2}  \\
 
\bottomrule
\end{tabular}
\vspace*{2ex}
\setlength{\belowcaptionskip}{-1.25em}
\caption{Sequential CIFAR10.
$\dagger$ denotes stationary. Chance accuracy is 10.0. 
}
\label{tab:cifar10}
\end{minipage}%
\hfill
\begin{minipage}[t]{.46\textwidth}
\centering
\vspace{0pt}
\begin{tabular}{l H c H c H c}
\toprule
& Model & Distill (LwF) & Model & ADI & Model & \textbf{\methodnameshort} \\
\midrule
Baseline (CIFAR10) & 4082 & 19.2 $\pm$ 0.3 & 2262 & 24.8 $\pm$ 0.9 & 3717 & \textbf{26.9 $\pm$ 1.1} \\
\midrule

Unit lag ($T=1$) & 6452 & 15.8 $\pm$ 2.6 & 6467 & 14.4 $\pm$ 1.6 & 4127 & \textbf{16.3 $\pm$ 1.8} \\

No distill & - & - & 2327 & 20.0 $\pm$ 0.8 & 2717 & \textbf{24.4 $\pm$ 1.7} \\
\bottomrule
\end{tabular}
\vspace*{2ex}
\caption{
Testing distillation based methods: \\unit lag and removing standard distillation. 
}
\label{tab:cifar10_2}
\vspace*{-0.5em}
    \includegraphics[width=0.315\textwidth]{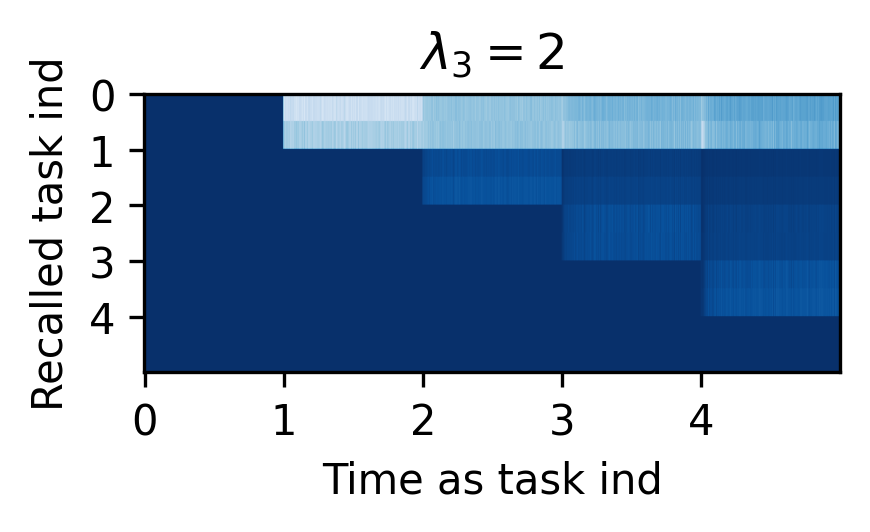}%
    \includegraphics[width=0.315\textwidth]{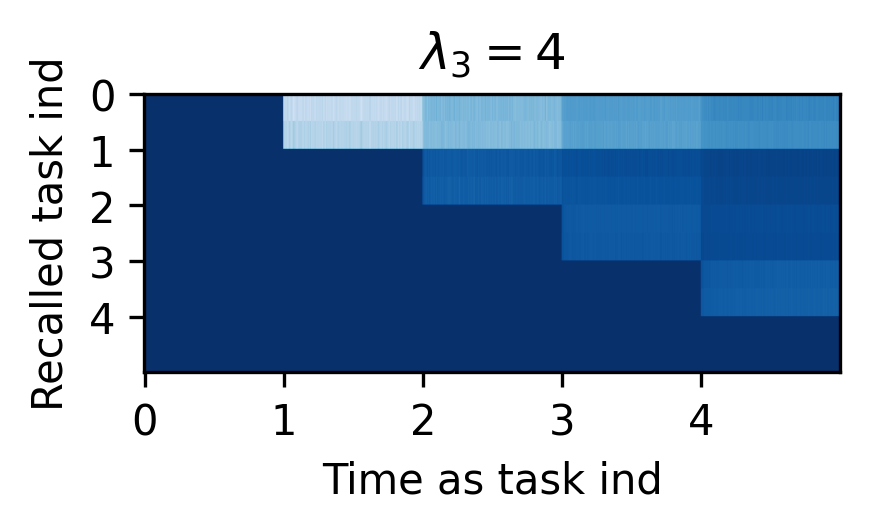}%
    \includegraphics[width=0.37\textwidth]{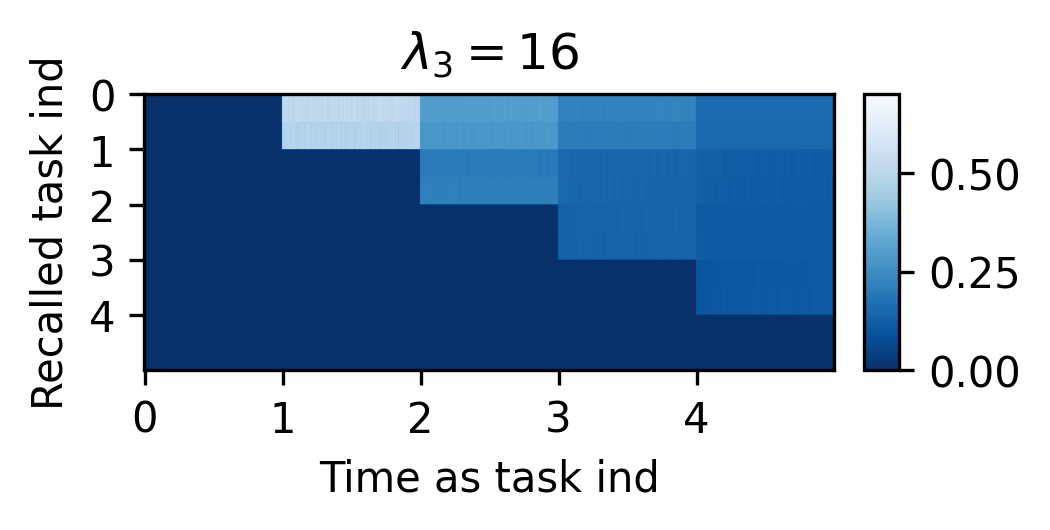}
\setlength{\belowcaptionskip}{-1.25em}
\captionof{figure}{History of recalled classes during training for different diversity $\lambda_3$, CIFAR10. 
}
\label{f:densities_combined}
\end{minipage}%
\end{table}

\subsection{Self-reinforcing loop between task performance and recall}\label{s:exp_empirical}
\begin{figure}[t]
\begin{subfigure}{.3\textwidth}
  \includegraphics[width=0.44\linewidth]{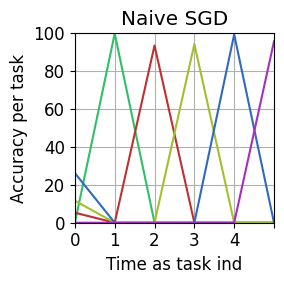}%
  \includegraphics[width=0.075\linewidth]{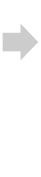}%
  \includegraphics[width=0.44\linewidth]{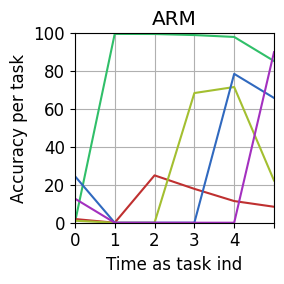}
  \caption{MNIST}
\end{subfigure}\hfill%
\begin{subfigure}{.3\textwidth}
  \includegraphics[width=0.44\linewidth]{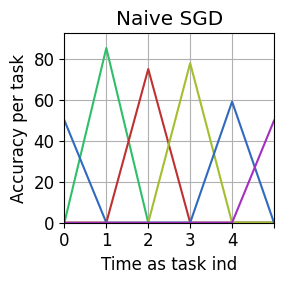}%
  \includegraphics[width=0.075\linewidth]{imgs/arrow.png}%
  \includegraphics[width=0.44\linewidth]{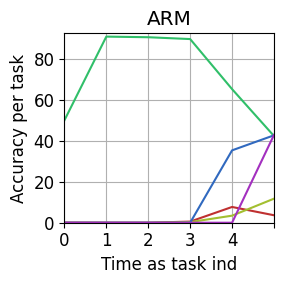}
  \caption{CIFAR10}
  \label{f:accuracy_cifar10}
\end{subfigure}\hfill%
\begin{subfigure}{.3\textwidth}
  \includegraphics[width=0.44\linewidth]{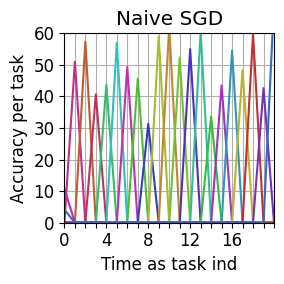}%
  \includegraphics[width=0.075\linewidth]{imgs/arrow.png}%
  \includegraphics[width=0.44\linewidth]{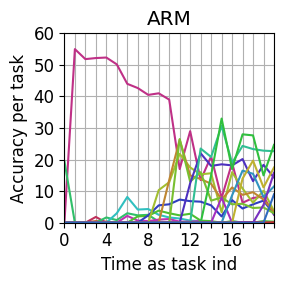}
  \caption{MiniImageNet}
  \label{f:accuracy_miniimagenet}
\end{subfigure}
\setlength{\belowcaptionskip}{-1.25em}
\caption{Recall causes characteristic changes to learning behavior. 
Tasks are retained beyond their training window and
positive backwards transfer~\citep{lopez2017gradient} is observed.
}
\label{f:accuracy}
\end{figure}

\methodnameshort reduces forgetting, observable in \cref{tab:cifar10} and \cref{f:accuracy} where accuracies remain elevated beyond the training window of the task.
In particular, we observed that the first task was simultaneously strongly recalled~(\cref{f:densities_combined}) and performed~(\cref{f:accuracy}) throughout training, far beyond its own training window.

The model tended to fixate on this strongly recalled task to the detriment of performance on other tasks, which were also not as strongly recalled.
The interplay between performance and recall is expected as recall in \methodnameshort depends entirely on the implicit memory of the tasks model; if a class is not retained by the tasks model, optimization cannot recover meaningful samples for that class from the weights.
This forms a parallel with the biological brain, where increased hippocampal replay leads to increased task performance~\citep{schuck2019sequential} and vice versa~\citep{mattar2018prioritized}.

\subsubsection{Abstract nature of recall}\label{s:abstract_recall}

Images from real $\mathcal{B}_X$ provide a favorable initialization in the optimization for maximally interfered $\mathcal{\hat{B}}_X$, since interference scales with the amount of feature overlap between them.
For MNIST, the recalled images are clearly reminiscent of the class of their associated target ~(\cref{f:mnist_prog}); for CIFAR10 and MiniImageNet, this is not the case, as the recalled images resemble their initialized images with added noise. 
However, we found that training on recalled samples produced gradients that were consistently more correlated with those produced from training on real images from the target class than real images from the originator class, with strongest correlations in the highest layers~(\cref{f:gradients}).
Thus we found that it was not necessary for input samples to empirically resemble a particular class for the  representation of that class to be reinforced, and it was the upper or abstract layers of the network where the most meaningful representational changes were induced by recall.
In the brain, mental imagery is thought not to propagate to the retina~\citep{pearson2015mental}, which has led to the idea of replay from intermediate levels in neural networks~\citep{vanbrain}.
However, this requires fixing parameters unaffected by replay, due to lack of protection from forgetting.
\methodnameshort demonstrates that replaying samples at input-output level, and thus protecting the trained network end-to-end, has effects that can nonetheless be characterized as abstract.

\subsubsection{Natural avoidance of unseen classes}\label{s:avoid_unseen}
\methodnameshort does not include any constraints on avoiding unseen classes.
Remarkably, we discover that it naturally does so, across diversity weights~(empty lower triangles, \cref{f:densities_combined}).
This is likely because unseen classes have relatively undiscriminative representations,
having thus far been trained not to activate, and are therefore de-prioritized in the optimization for samples with diverging outputs.

\vspace*{-0.9em}
\section{Conclusion}
\vspace*{-0.8em}
Avoiding catastrophic forgetting in artificial neural networks naturally gives rise to recall mechanisms reminiscent of human cognition.
Internally generated, conditional replay is a promising approach that reduces catastrophic forgetting, memory complexity, and outperforms other forms of generative replay on natural images.

\section*{Broader Impact}\vspace*{-0.5em}
This work deals with artificial neural networks internally generating imagined data.
Research at the intersection of machine and biological intelligence has the potential to give back to neuroscience and further the understanding of our own cognition, which is a deeply compelling objective.
Removing the need to explicitly store buffered data also helps algorithms overcome data protection requirements.

\bibliographystyle{plainnat}
\bibliography{refs.bib}
\newpage
\appendix
\section{Appendix}

\subsection{Background}\label{s:related}

In this section we discuss four underlying principles of our method.

\subsubsection{Distributed representations}\label{s:distributed}

\methodnameshort trains a single neural network for all tasks, avoiding tables and task-specific parameters, as these factors introduce enforced non-distributedness.
Neural networks generally implement distributed representations, meaning inputs are represented by overlapping patterns.
Distributedness allows for high memory efficiency~\citep{bengio2013representation}; for example, $n$ bits stores $2^n$ patterns if overlap is allowed and $n$ if not. 
Overlap also means representations are shared between and thus optimized across tasks, minimising redundancy from duplication.
Behaviour generalises to patterns not present in training, which is not true of non-distributed representations such as lookup tables~\citep{french1991using}. 

Tables and buffers~\citep{graves2014neural, mnih2013playing, chaudhry2019continual} are examples of non-distributed representations, having separated cells that allow for atomic updates without interference, but sacrificing on memory efficiency, transfer and generalisation~\citep{french1991using, bengio2013representation}.
Rather than allowing the degree of distributedness to be naturally determined from optimization for an objective, as seen in neural network training, non-distributedness is enforced a priori.
Dynamic architectures~\citep{rusu2016progressive} also exploit separation, as previous parameters can be isolated as tasks change, resulting in high memory complexity with respect to the number of tasks.

Data representation in the brain is distributed, where the number of stimuli that can be represented increases exponentially with the number of neurons~\citep{rolls1997representational}. 
Distributedness varies, with for example regions implicated in fast-learning episodic memory exhibiting greater pattern separation~\citep{kumaran2016learning}, but on a continuous spectrum 
rather than with the hard difference seen between neural network and tabular representations.

\subsubsection{One model for inference and recall}\label{s:related_rehearsal}

In \methodnameshort the invertibility of the tasks model, with its implicit memory of past training samples, is used to generate recalled samples. 
Buffers or models trained for sample generation are not used.

In contrast, replay methods typically require extra memory explicitly dedicated to the auxiliary task of sample memorisation, either in the form of buffers~\citep{chaudhry2019continual,aljundi2019online, aljundi2019gradient, riemer2018learning, rebuffi2017icarl, rolnick2019experience} or dedicated generators~\citep{shin2017continual, kamra2017deep, lavda2018continual, lesort2019marginal, atkinson2018pseudo, kemker2017fearnet}. Some parameter-level constraints (i.e. non-replay) also make use of buffers~\citep{lopez2017gradient, chaudhry2018efficient}. Buffers being non-distributed are relatively memory inefficient. Generator networks have so far proved difficult to train given only non-stationary data in class-incremental settings~\citep{lesort2019generative, aljundi2019online}.
Notable exceptions with internal generation, with key differences to our method highlighted, include Adaptive DeepInversion~\citep{yin2019dreaming} (generates full seen data distribution, so classes are arbitrarily pre-selected and representativeness of batch statistics is enforced), and Replay-through-Feedback~\citep{vanbrain, van2018generative} (tasks model is overlaid with encoder of VAE, thus shares decoder parameters in symmetric case; again, sampling is intended to represent full distribution).

The brain also appears to share memory for inference and recall; namely the same neurons activated during experiences are re-activated during their recall~\citep{gelbard2008internally, carr2011hippocampal}. 

\subsubsection{Functional rather than parameter-level constraints}
A neural network can have multiple instantiations with different parameter values that execute the same overall behavior~\citep{williamson1995existence}. Thus explicit constraints aimed at preventing changes in function should be specified at the functional level, i.e. input-output, rather than at parameter level, as noted in~\cite{ramapuram2020lifelong}.
Note the two are related as functional-level constraints are implemented as parameter-level constraints in parametric models, and parameter-level constraints also constrain function; the difference lies in that a functional-level constraint can be implemented as multiple parameter-level constraints but not vice-versa.
The stringency of parameter level regularization may explain its underperformance compared to replay~\citep{van2019three, chaudhry2019continual}. 
\methodnameshort offers the benefit of being data-free, like some parameter level regularizers, while being a replay method and therefore constraining at functional level.

Like \methodnameshort, LwF~\citep{li2017learning} is an example of imposing functional constraints without needing to explicitly store the seen data distribution. 
Unlike \methodnameshort, the replay samples are taken from the current batch without optimization.
Similarly, \methodnameshort is a form of distillation~\citep{hinton2015distilling}.

Parameter-level methods include EWC~\citep{kirkpatrick2017overcoming}, SI~\citep{zenke2017continual}, GEM~\citep{lopez2017gradient}, A-GEM~\citep{chaudhry2018efficient}, Meta Continual Learning~\citep{vuorio2018meta}, IMM~\citep{lee2017overcoming}, Memory Aware Synapses~\citep{aljundi2018memory}.
These methods generally explictly anchor either the values or gradients of parameters from excessively changing. 

From an evolutionary perspective, it is overall behavior that is directly implicated for survival, as opposed to the local behaviour of specific neurons.
For example, a major influence on representations in the brain is optimization for rewards from tasks~\citep{miller2001integrative}.

\subsubsection{Conditionality of recall and top-down modulation}\label{s:conditionality}
In \methodnameshort, recalled samples are optimized given model snapshots separated by training on the current real batch, and are thus conditioned on the latter. 
Thus we avoid the replay-all-seen-tasks paradigm that is inefficient and unsustainable for large numbers of tasks. 
Furthermore, by generating subsequent training inputs by backpropagation - analogous to a forward pass in the opposing direction - \methodnameshort exhibits literal top-down modulation of subsequent activations and learning within the tasks network.

Few generative replay methods~(\cref{s:related_rehearsal}) consider conditionality on the temporally local environment. Rather, most aim to reproduce the full seen data distribution, with the idea that training a network with replay simulates stationary training when replay is representative~\citep{shin2017continual}. Thus randomization is used to select from buffers or choose codes or classes for decoding samples.
In contrast, \cite{aljundi2019online} is a replay method that selectively replays based on prediction change caused by training on the current real batch.
\methodnameshort provides a theoretical argument for why this is reasonable, as well as using internal generation instead of external buffers or generators.

Human recall is also a response conditioned on the environment as opposed to uniformly sampled across all knowledge. 
Memory retrieval is conditioned on current sensory input~\citep{carr2011hippocampal}.
Top-down modulation is a form of conditionality where activations representing higher-level or downstream concepts influence the activations of lower-level or upstream representations, reversing the bottom-up processing direction~\citep{miller2001integrative}.
It is believed that many of the cognitive capabilities associated with the prefrontal cortex, namely various forms of reasoning, depend on its top-down attentional modulation of other areas, conditioned on the current goal~\citep{miller2001integrative, russindeep}.
Note that the backpropagation algorithm itself can be seen as a form of attentional modulation of lower-level representations based on higher-level activations.
However, unlike the brain, which is highly recurrent~\citep{kumaran2016learning}, this top-down modulation is truncated in that it does not yield subsequent forward pass activations nor additional learning steps. 
In \methodnameshort, this is corrected as an initial learning step produces recalled inputs, yielding subsequent activations and an additional learning step for reinforcing these activations.
This allows for a chain of inference induced by training on the environment that is longer than 1 pass over memory.

\begin{center}
\begin{minipage}[t]{0.88\textwidth}
\DontPrintSemicolon
\fontsize{8}{8}\selectfont
\begin{algorithm}[H]
\fontsize{8}{8}\selectfont
\SetAlgoLined
 Require: randomly initialized $f_\theta$, data $D$, lag $T$, batch sizes $N$ and $M$, steps $S$, rates $\eta_{0,1}$ and $\lambda_{0 \dots 6}$. \;
 $\theta' = \theta$\;
 \For{$t \in [0, |D|)$}{
   $\mathcal{B} = D_t$\;
   $\theta' = \theta' - \eta_0 \nabla_{\theta'} \frac{1}{N} \sum_{(x, y) \in \mathcal{B}} \operatorname{L}(f_{\theta'}(x), y)$  \hspace{8.5em} $\triangleright$ \hspace{0.25em} inference and learning on $D$ \;
   $\mathcal{\hat{B}}_X = \mathcal{B}_X[i_{0..M-1}], i \sim U(0, N-1)$\;
     \For{$s \in [0, S)$}{
       $\mathcal{\hat{B}}_X = \mathcal{\hat{B}}_X + \eta_1 \nabla_{\mathcal{\hat{B}}_X} \operatorname{recall}(\mathcal{\hat{B}}_X;\mathcal{B}, f_\theta, f_{\theta'})$ 
       \hspace{6.17em} $\triangleright$ \hspace{0.25em} infer recalled samples
       \;
     }
     $\theta' = \theta' - \eta_0 \nabla_{\theta'} \frac{1}{M + N} \sum_{\hat{x} \in \mathcal{\hat{B}}_X \bigcup \mathcal{B}_X} \lambda_0 \operatorname{L}(f_{\theta'}(\hat{x}), f_{\theta}(\hat{x})) $
     \hspace{1.85em} $\triangleright$ \hspace{0.25em} learn on recalled samples
     \label{algo:distill}
     \;
     
     \If{$(t+1)\text{ }\%\text{ }T = 0$}{
     $\theta = \theta'$\;
     }
}
 \caption{\methodname with distillation}\label{algo}
\end{algorithm}
\end{minipage}
\end{center}

\subsection{Datasets}

\begin{table}[h]
    \centering
    \fontsize{7}{7}\selectfont
    \begin{tabular}{l  c c c c c c c}
    \toprule
    Dataset & Tasks & Classes & Classes per task & \#Train & \#Test & \#Val & Batch size $|\mathcal{B}|$  \\
    \midrule
    CIFAR10 & 5 & 10 & 2 & 47.5K & 10K & 2.5K & 10 \\
    MiniImageNet & 20 & 100 & 5 & 45.6K & 12K & 2.4K & 10 \\
    MNIST & 5 & 10 & 2 & 5K & 10K & 3K & 10 \\
    \bottomrule
    \end{tabular}
    \setlength{\abovecaptionskip}{1em}
    \setlength{\belowcaptionskip}{-0.5em}
    \caption{Dataset statistics.}\label{t:datasets}
\end{table}%
Datasets are summarized in~\cref{t:datasets} and follow~\cite{aljundi2019online}.
Pre-processing was not used except resizing MiniImageNet images to 84x84 as standard.

\subsection{Evaluation}
Average accuracy and forgetting (average drop in accuracy) were measured on a held-out test set at the end of training on all tasks, as defined in \cite{chaudhry2019continual}.

\textbf{Average accuracy.} Let $a_{i, j}$ denote performance on the test set of task $j$ after training on task $i$. The average accuracy after task $T$ is:
{\small
\begin{equation} \label{eq: avg_acc}
    A_{T} = \frac{1}{T} \sum_{j=1}^T a_{T, j}.
\end{equation}
}%
\textbf{Forgetting.} Average forgetting after task $T$ is:
{\small
\begin{align}
    F_T &= \frac{1}{T-1} \sum_{j=1}^{T-1} f_j^T \label{eq:fgt} \\
    \text{where } f_j^i &= \max_{l \in \{1,\cdots,i-1\}} a_{l, j} - a_{i, j}.
\end{align}
}%

\subsection{Architectures and memory computations}
Following~\cite{aljundi2019online}, ResNet18 was used for CIFAR10 and MiniImageNet, and a multi-layer perceptron with two hidden layers was used for MNIST.
The number of parameters for each is given in~\cref{tab:cifar10,tab:miniimagenet,tab:mnist}.
To compute auxiliary memory usage, we assumed 3 bytes (RGB) or 1 byte (grayscale) per pixel for sample memory, and 4 bytes (single precision floating point) per parameter for model memory.

\subsection{Hyperparameters}
\begin{table}[h]
    \fontsize{7}{7}\selectfont
    \centering
    \begin{tabular}{l  c c c c  c c c c c c  c}
    \toprule
    Dataset & $\eta_0$ & $\eta_1$ & $\lambda_0$ : $\mathcal{\hat{B}}_X$ & $\lambda_0$: $\mathcal{B}_X$ & $\lambda_1$ & $\lambda_2$ & $\lambda_3$ & $\lambda_4$ & $\lambda_5$ & $\lambda_6$ & $S$ \\
    \midrule
    MNIST  & 0.05 & 25.0 & 1.0 & 1.0 & 1.0 & 0.1 & 16.0 & 0.1 & 1.0 & 1.0 & 10 \\
    CIFAR10 & 0.01 & 10.0 & 1.0 & 1.0 & 1.0 & 0.1 & 16.0 & 0.1 & 1.0 & 1.0 & 10 \\
    MiniImageNet & 0.01 & 10.0 & 1.0 & 2.0 & 1.0 & 0.1 & 1.0 & 1.0 & 1.0 & 1.0 & 10 \\
    \bottomrule
    \end{tabular}
    \setlength{\abovecaptionskip}{1em}
    \setlength{\belowcaptionskip}{-0.5em}
    \caption{Hyperparameters.}
    \label{tab:hyperparams}
\end{table}
Hyperparameter values were selected based on performance on a held-out validation set.
A grid search was conducted on CIFAR10 for $\lambda_{1 \dots 4}$ and other values were finalized by ablation.
Hyperparameters for \methodnameshort experiments in~\cref{tab:cifar10,tab:miniimagenet,tab:mnist} are given in~\cref{tab:hyperparams}.
Experiments in~\cref{tab:cifar10_2} use the same values except $\lambda_4=1.0$ for no distill, and an additional weight of 8.0 on $\operatorname{D}$~(\cref{e:recall}) for unit lag, as we found it was beneficial to increase the emphasis on maximal divergence in this case.
Hyperparameter values for ADI and LwF were determined in the same manner.
The implementation of \cite{aljundi2019online} was used for ER and GEN experiments.

\subsection{Code}
The implementation can be found at \url{www.github.com/xu-ji/ARM}.

\newpage

\subsection{Experiments}

\subsubsection{Protocol}
Tasks are sequential and formed by partitioning training classes equally.
Output classes are unconstrained throughout training and evaluation (single head,~\cite{farquhar2018towards}).
Batch size is kept constant as it affects effective task training length.
As in~\cite{aljundi2019online}, replay begins from the end of the 1st task.
The update lag $T$ is set as the number of training iterations per task (equivalently, updating when a batch contains novel classes)~\citep{aljundi2019online} but we also test unit lag.
Hyperparameters are selected using validation set, evaluation uses test set, and our experiments are repeated 5 times.
Hyperparameters and ablations are given in the Appendix.

Few generative methods target class-incremental learning on natural images, which is very difficult in our online single-pass setting.
Other works generally test online generation on digits (MIR~\citep{aljundi2019online}, GEN~\citep{shin2017continual}), pre-processed features~\citep{kemker2017fearnet}, or avoid class-incremental learning by viewing entire datasets as a single task (ADI~\citep{yin2019dreaming}).

\subsubsection{Tunable diversity of recall}
\begin{figure}[t]
\centering
\includegraphics[width=0.5\linewidth]{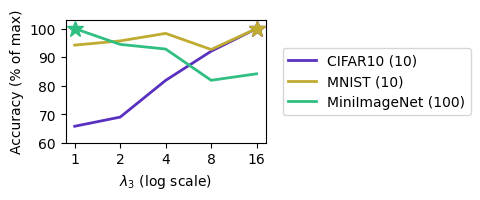}
\setlength{\belowcaptionskip}{-1.5em}
\captionof{figure}{
Optimal diversity weight ($\star$) is negatively correlated with the number of classes.}
\label{f:accuracy_with_divs}
\end{figure}

We use almost identical hyperparameters across all three datasets, as in most cases the same values work well.
Varying diversity via $\lambda_3$ biases recall towards sparse or dense~(\cref{f:densities_combined}).
High $\lambda_3=16$ is optimal for MNIST and CIFAR10
while low $\lambda_3=1$ is optimal for MiniImageNet~(\cref{f:accuracy_with_divs}), which benefits more from selectivity as the space of possible choices is much larger.
With a low diversity weight on CIFAR10, we observed a natural prioritization of recent classes~(\cref{f:densities_combined}, left).
This is likely due to reduced but present forgetting causing a decrease in the discriminativeness of older class representations relative to recent classes, resulting in the latter being highlighted in the optimization for samples with diverging outputs.
In the brain, replay is also most prevalent immediately following an experience, and decays with time~\citep{kudrimoti1999reactivation,karlsson2009awake}.

\begin{table}[b]
\fontsize{6}{6}\selectfont
\setlength{\tabcolsep}{2pt}
\begin{minipage}[t]{.49\textwidth}
\vspace{0pt}
\begin{tabular}{l H c c c c c c c}
\toprule
Method & Models & \multicolumn{2}{c}{+Sample mem.} & \multicolumn{2}{c}{+Model mem.} & $M$ & Accuracy & Forgetting \\
& & \#images & MB & \#params & MB & & & \\
\midrule

Naive SGD$\dagger$ & 4527 & 0 & 0 & 0 & 0 & 0 & 25.2 $\pm$ 4.7 & - \\

\midrule

ER & - & 500 & 10.6 & 0 & 0 & 10 & 5.60 $\pm$ 0.4 & 55.2 $\pm$ 1.2 \\

ER-MIR & - & 500 & 10.6 & 0 & 0 & 10 & 5.40 $\pm$ 0.4 & 55.8 $\pm$ 0.9 \\

ER & - & 100 & 2.12 & 0 & 0 & 10 & 4.20 $\pm$ 0.4 & 58.6 $\pm$ 0.9 \\

ER-MIR & - & 100 & 2.12 & 0 & 0 & 10 & 4.00 $\pm$ 0.0 & 59.2 $\pm$ 0.9 \\

\midrule

Naive SGD & 4557 & 0 & 0 & 0 & 0 & 0 & 3.80 $\pm$ 0.2 & 51.2 $\pm$ 2.0  \\

GEN & (13) & 0 & 0 & 27.1M & 108 & 10 & 2.33 $\pm$ 0.3 & 34.3 $\pm$ 0.7 \\

GEN-MIR & (9) & 0 & 0 & 27.1M & 108 & 20 & 2.00 $\pm$ 0.0 & 35.3 $\pm$ 0.5 \\

Distill (LwF) & 6092 & 0 & 0 & 1.16M & 4.63 & 100 & 4.89 $\pm$ 0.2 & 41.6 $\pm$ 1.6  \\

ADI & 6042 & 0 & 0 & 1.16M & 4.63 & 100 & 2.99 $\pm$ 0.6 & 8.11 $\pm$ 2.1 \\

\textbf{\methodnameshort} & 4821 & \textbf{0} & \textbf{0} & \textbf{1.16M} & \textbf{4.63} & \textbf{100} & \textbf{5.53 $\pm$ 0.8} & \textbf{11.9 $\pm$ 2.7} \\
 
\bottomrule
\end{tabular}
\vspace*{2ex}
\caption{Sequential MiniImageNet.
$\dagger$ denotes stationary.
3 task iterations made, as with~\cite{aljundi2019online}; stationary uses 3 epochs. Chance accuracy is 1.0.}
\label{tab:miniimagenet}
\end{minipage}%
\hfill
\begin{minipage}[t]{.48\textwidth}
\vspace{0pt}
\begin{tabular}{l H c c c c c c c}
\toprule
Method & Models & \multicolumn{2}{c}{+Sample mem.} & \multicolumn{2}{c}{+Model mem.} & $M$ & Accuracy & Forgetting \\
& & \#images & MB & \#params & MB & & & \\
\midrule

Naive SGD$\dagger$ & 4947 & 0 & 0 & 0 & 0 & 0 & 87.1 $\pm$ 0.6 & - \\

\midrule
ER & - & 50 & 0.04 & 0 & 0 & 10 & 62.8 $\pm$ 3.1 & 42.0 $\pm$ 3.7 \\

ER-MIR & - & 50 & 0.04 & 0 & 0 & 10 &  63.8 $\pm$ 4.6 & 40.6 $\pm$ 5.9 \\

ER & - & 25 & 0.02 & 0 & 0 & 10 & 51.6 $\pm$ 2.7 & 57.0 $\pm$ 3.3 \\

ER-MIR & - & 25 & 0.02 & 0 & 0 & 10 & 51.6 $\pm$ 2.6 & 56.4 $\pm$ 3.3 \\

\midrule
Naive SGD & 4967 & 0 & 0 & 0 & 0 & 0 & 18.8 $\pm$ 0.5 & 95.6 $\pm$ 2.7 \\

GEN & - & 0 & 0 & 1.14M & 4.58 & 40 & 79.3 $\pm$ 0.6 & 19.5 $\pm$ 0.8\\

GEN-MIR & - & 0 & 0 & 1.08M & 4.31 & 100 & 82.1 $\pm$ 0.3 & 17.0 $\pm$ 0.4\\

Distill (LwF) & 6102 & 0 & 0 & 478K & 1.91 & 10 & 33.3 $\pm$ 2.5 & 58.0 $\pm$ 1.7  \\

ADI & 6232 & 0 & 0 & 478K & 1.91 & 10 & 55.4 $\pm$ 2.6 & 11.5 $\pm$ 5.0  \\

\textbf{\methodnameshort} & 6579 & \textbf{0} & \textbf{0} & \textbf{478K} & \textbf{1.91} & \textbf{10} & \textbf{56.3 $\pm$ 2.6} & \textbf{21.8 $\pm$ 1.6} \\

\bottomrule
\end{tabular}
\vspace*{2ex}
\caption{Sequential MNIST. 
$\dagger$ denotes stationary.
Chance accuracy is 10.0.}
\label{tab:mnist}
\end{minipage}

\end{table}

\subsubsection{Benefits of optimized selection and distributed memory}
We hypothesized that the performance boost from allowing classes to be selective (\methodnameshort, *-MIR) as opposed to random (ADI, ER, GEN) would be most obvious with a large number of classes. The benefit of distributed (\methodnameshort, ADI, GEN-*) over non-distributed (ER-*) data stores should also be more obvious due to the lower memory complexity of distributed representations~(\cref{s:distributed}).
Unfortunately, datasets with more classes are also more difficult in general, with lower baseline or chance performance.
However, we were able to verify both these hypotheses on the 100 classes of MiniImageNet~(\cref{tab:miniimagenet}).
The 2.5\% accuracy boost from \methodnameshort compared to ADI equates to the learning of 2.5 additional classes or 50\% of a task.
In contrast, the boost on MNIST equates to only a tenth of a class or 5\% of a task~(\cref{tab:mnist}).
On MiniImageNet, storing the old model costs 4.6MB but achieves comparable performance to storing 500 images at a cost of 10.6MB.

\subsubsection{Lower memory complexity than external generative replay}
External generative replay methods GEN and GEN-MIR
perform worse than \methodnameshort on natural images~(\cref{tab:cifar10,tab:miniimagenet}), despite incurring greater memory costs. This is because 3 additional networks must be stored: old and new versions of the generator network and an old version of the tasks network, whereas \methodnameshort uses only the latter.
In particular, GEN and GEN-MIR struggled on MiniImageNet and underperformed the naive SGD baseline.
These results support the hypothesis that for online continual learning on datasets of moderate to high complexity, obtaining meaningful samples from the existing tasks model is easier than training a separate generator.

\subsubsection{Ablation}

\begin{table}[h]
    \begin{varwidth}{0.45\textwidth}
    \fontsize{6}{6}\selectfont
  \setlength{\tabcolsep}{3pt}
    \centering
    \begin{tabular}{l H c c}
    \toprule
    & & Accuracy & Forgetting \\
    \midrule
    Baseline (CIFAR10) & 3717 & 26.4 $\pm$ 1.2 & 8.07 $\pm$ 5.2 \\
    \midrule
    $\lambda_1 = 0, \lambda_2 = 0$ & 3982 & 22.8 $\pm$ 1.8 & 29.1 $\pm$ 2.0\\
    $\lambda_3 = 0$ & 6562 & 15.3 $\pm$ 1.3 & 4.17 $\pm$ 0.9 \\
    $\lambda_4 = 0$ & 3977 & 25.7 $\pm$ 2.0 & 6.69 $\pm$ 5.4 \\
    $\lambda_5 = 0$ & 3967 & 25.8 $\pm$ 1.2 & 6.95 $\pm$ 5.1 \\
    $\lambda_6 = 0$ & 3972 & 26.4 $\pm$ 1.7 & 6.36 $\pm$ 3.7 \\
    \midrule
    $M = 150$ (+50) & 6478 & 26.3 $\pm$ 0.9 & 16.5 $\pm$ 5.5 \\
    $M = 50$ (-50) & 5922 & 24.4 $\pm$ 1.2 & 5.71 $\pm$ 4.5 \\
    $S = 20$ (doubled) & 3957 & 25.1 $\pm$ 1.1 & 8.11 $\pm$ 7.2 \\
    $S = 5$ (halved) & 3962 & 24.6 $\pm$ 1.5 & 13.5 $\pm$ 8.4 \\
    \midrule
    Cross-entropy as $D$ & 3617 & 24.6 $\pm$ 1.2 & 24.9 $\pm$ 5.2 \\
    Random noise init $\mathcal{\hat{B}}_X$ & 4077 & 16.0 $\pm$ 1.4 & 4.00 $\pm$ 1.1 \\
    Recall 2x per $t$ & 6502 & 17.5 $\pm$ 0.6 & 0.57 $\pm$ 0.5 \\
    Recall 4x per $t$ & 6507 & 17.4 $\pm$ 1.0 & 0.44 $\pm$ 0.9 \\
    \bottomrule
    \end{tabular}
    \setlength{\abovecaptionskip}{1em}
    \caption{Ablation study on CIFAR10. Each result makes a change from the baseline.}
    \label{tab:ablation}
\end{varwidth}%
\hfill
\begin{minipage}{0.5\textwidth}
\centering
\begin{subfigure}{.6\textwidth}
\includegraphics[width=\linewidth]{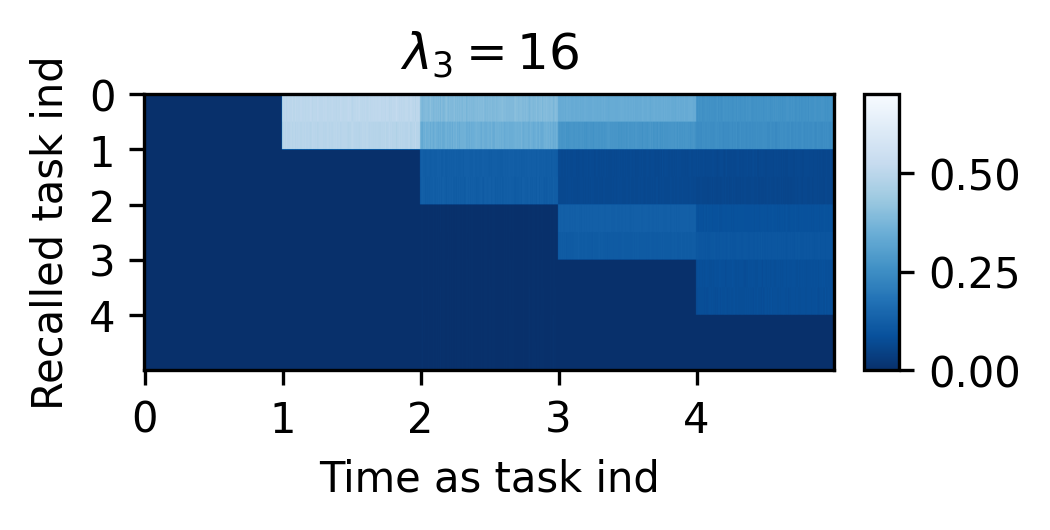}
\caption{}
\end{subfigure}\hspace*{1em}%
\begin{subfigure}{.3\textwidth}
\includegraphics[width=\linewidth]{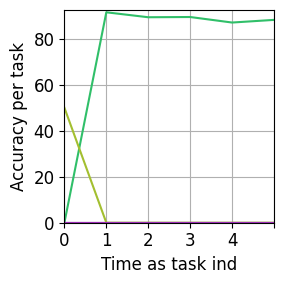}
\caption{}
\end{subfigure}
\label{f:suppressed_warmup}
\captionof{figure}{
CIFAR10 with 2 recalls per timestep $t$. Contrast with \cref{f:densities_combined} right, and \cref{f:accuracy_cifar10}.
}
\label{f:cifar10_more_theta_steps}
\end{minipage}
\end{table}

The contributions of different \methodnameshort implementation details were measued with ablation~(\cref{tab:ablation}).
Among the least important was image regularization; L2 norm ($\lambda_5$) and TV minimization ($\lambda_6$)  were added mainly for fairness with ADI, which also uses them. 
\methodnameshort still outperformed ADI without them.
The most important factors were initializing recalled samples from the real batch instead of random noise, and using entropy weight ($\lambda_3$) to minimize duplication.

$M$ is used to denote auxiliary replay batch size. Note the material ``size'' for replay is not $M$, but additional persistent memory size. 
Deliberate subsampling of the buffer ($M$ less than size of buffer) is used in~\cite{aljundi2019online}, which prevents identical replay batches across training iterations. This does not apply to \methodnameshort as replay images are not sampled from a fixed buffer.
On CIFAR10, we found an increase in performance up to $M=100$ and diminishing returns thereafter.

Our implementation does not provide a formal guarantee of not forgetting as we only recall once per training iteration, taking a single learning step towards minimizing divergence~(\cref{algo}, line 9). 
We tested increasing the number of training steps from recall, looping from line 9 to line 6 in~\cref{algo}.
This was found to lead to very strong recall of the first task throughout the training sequence, despite high $\lambda_3$~(\cref{f:cifar10_more_theta_steps}), and decreased overall performance~(\cref{tab:ablation}).
\methodnameshort typically demonstrates a warm-up phase with fixation on the first task gradually reducing to accommodate other tasks~(\cref{f:accuracy}), which was suppressed in this case~(\cref{f:suppressed_warmup}).
This result underlines the need to consider both forgetting and accuracy, as a near eradication of forgetting ($<1\%$ average drop in accuracy, \cref{tab:ablation}) is not necessarily indicative of optimal overall performance.
In summary, balanced training with one recall batch per real batch was found to perform best in our experiments.

\newpage
\subsubsection{Gradients study}

\begin{figure}[h]
  \centering
  \fontsize{6}{6}\selectfont
  \setlength{\tabcolsep}{3pt}
  \setlength{\belowcaptionskip}{-0.5em}
  \begin{subfigure}{.33\textwidth}
  \includegraphics[width=\textwidth]{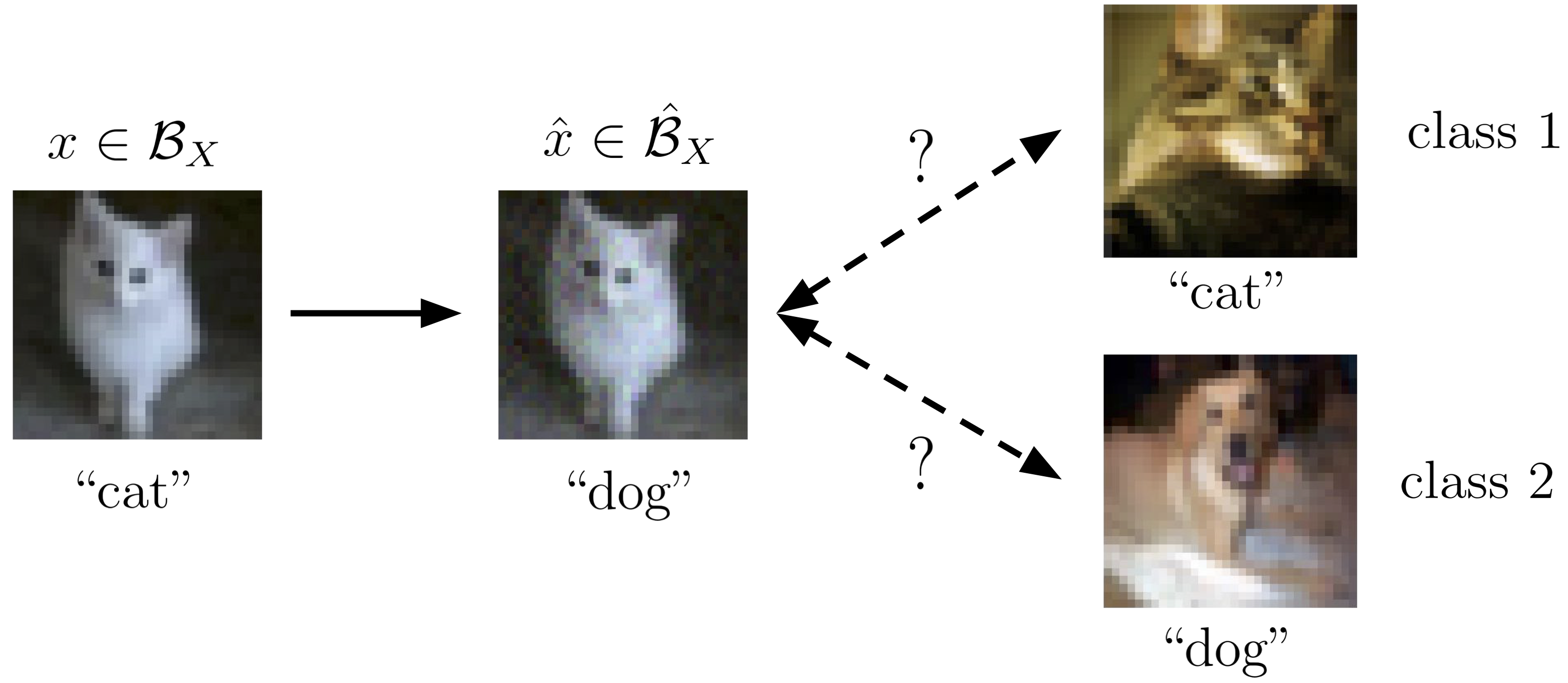}
  \label{f:gradients1}
  \end{subfigure}%
  \begin{subtable}{0.33\textwidth}
  \centering
    \begin{tabular}{l c c}
        \toprule
        Layer & Corr. class 1 & Corr. class 2 \\
        \midrule
        Conv 1 & 0.034 $\pm$ 0.281 & 0.205 $\pm$ 0.254 \\
        
        Conv 2  & 0.036 $\pm$ 0.262 & 0.259 $\pm$ 0.214 \\
        
        Conv 3  & 0.048 $\pm$ 0.315 & 0.344 $\pm$ 0.234 \\
        
        Conv 4  & 0.036 $\pm$ 0.362 & 0.366$\pm$ 0.277 \\
        
        Conv 5  & 0.012 $\pm$ 0.308 & 0.419 $\pm$ 0.210 \\
        
        FC  & 0.002 $\pm$ 0.365 & 0.687 $\pm$ 0.143 \\
        
        \midrule
        All & 0.032 $\pm$ 0.278 & 0.355 $\pm$ 0.196 \\
        \bottomrule
    \end{tabular}
    \caption{Hard targets (one-hot) $\mathcal{\hat{B}}_Y$}
    \label{t:gradients2}
  \end{subtable}%
  \begin{subtable}{0.33\textwidth}
    \centering
    \begin{tabular}{l c c}
        \toprule
        Layer & Corr. class 1 & Corr. class 2 \\
        \midrule
        Conv 1 & 0.016 $\pm$ 0.270 & 0.084 $\pm$ 0.275 \\
        
        Conv 2  & 0.001 $\pm$ 0.263 & 0.080 $\pm$ 0.261 \\
        
        Conv 3  & -0.017 $\pm$ 0.312 & 0.060 $\pm$ 0.323 \\
        
        Conv 4  & -0.046 $\pm$ 0.364 & 0.040 $\pm$ 0.388 \\
        
        Conv 5  & -0.052 $\pm$ 0.313 & 0.096 $\pm$ 0.332 \\
        
        FC  & -0.096 $\pm$ 0.351 & 0.137 $\pm$ 0.395 \\
        
        \midrule
        All & -0.030 $\pm$ 0.279 & 0.071 $\pm$ 0.298 \\
        \bottomrule
    \end{tabular}
    \caption{Soft targets $\mathcal{\hat{B}}_Y$}
    \label{t:gradients3}
  \end{subtable}
\setlength{\belowcaptionskip}{-1em}
\caption{
Despite $\hat{x}$ resembling class 1 at pixel level~(a), training on $\hat{x}$ produces gradients more correlated with training on real images of class 2 than real images of class 1, with correlation being strongest in the highest layers of the network. 
Hard targets~(b) are considered as well as soft~(c) for fairness, as real samples use hard targets.
Results shown on CIFAR10 using 1K random samples of $\hat{x}$.
}
\label{f:gradients}
\end{figure}
As discussed in \cref{s:abstract_recall}, training on recalled samples produced gradients that were consistently more correlated with those produced from training on real images from the target class than real images from the originator class, in particular in the highest layers~(\cref{f:gradients}), even when recalled samples resembled the originator class at pixel level to the human eye. 
This was observed across all datasets and architectures, including both convolutional ResNet and multilayer perceptrons.
Gradients within each layer were normalized for correlation, so this behavior is not attributable to varying magnitudes. Rather, it indicates that similarities in output targets have greater influence than similarities in input space in the computation of gradients, supporting the argument that recalled targets are material and should be optimized rather than arbitrary~(\cref{s:conditionality}).
In short, recalled samples can be deceptive at pixel level and require an analysis of gradients to understand the relation to training on real images.

Computing correlations for~\cref{f:gradients,t:gradients_mnist,t:gradients_miniimagenet} involved gradients computed from 1K real samples and 1K recalled samples per dataset.
The underlying models were taken from the end of training and parameters were fixed.
Gradients for each parameter block (i.e. weights and biases) were linearized and normalized.
2 dot products were computed per parameter block and recalled sample, between the gradients induced by recall and the gradients induced by training on each of 2 randomly selected real samples belonging to the recalled sample's originator class (class 1) and target class (class 2) respectively.
Results were collected per layer and averaged over recalled samples and parameter blocks.

\begin{table}[h]
  \centering
  \fontsize{6}{6}\selectfont
  \setlength{\tabcolsep}{3pt}
  \begin{subtable}[h]{0.45\textwidth}
    \centering
    \begin{tabular}{l c c}
        \toprule
        Layer & Corr. class 1 & Corr. class 2 \\
        \midrule
        Conv 1 & 0.061 $\pm$ 0.316 & 0.246 $\pm$ 0.287 \\
        
        Conv 2 & 0.032 $\pm$ 0.274 & 0.282 $\pm$ 0.241 \\
        
        Conv 3 & 0.028 $\pm$ 0.326 & 0.414 $\pm$ 0.258 \\
        
        Conv 4 & -0.031 $\pm$ 0.264 & 0.474 $\pm$ 0.198 \\
        
        Conv 5 & -0.181 $\pm$ 0.195 & 0.556 $\pm$ 0.142 \\
        
        FC & -0.014 $\pm$ 0.084 & 0.726 $\pm$ 0.084 \\
        
        \midrule
        All & -0.036 $\pm$ 0.218 & 0.439 $\pm$ 0.177 \\
        \bottomrule
    \end{tabular}
    \caption{Hard targets (one-hot) $\mathcal{\hat{B}}_Y$}
    \label{t:gradients_miniimagenet1}
  \end{subtable}
  \begin{subtable}[h]{0.45\textwidth}
  \centering
    \begin{tabular}{l c c}
        \toprule
        Layer & Corr. class 1 & Corr. class 2 \\
        \midrule
        Conv 1 & 0.065 $\pm$ 0.319 & 0.212 $\pm$ 0.302 \\
        
        Conv 2 & 0.022 $\pm$ 0.273 & 0.262 $\pm$ 0.259 \\
        
        Conv 3 & 0.034 $\pm$ 0.315 & 0.381 $\pm$ 0.273 \\
        
        Conv 4 & -0.040 $\pm$ 0.247 & 0.433 $\pm$ 0.216 \\
        
        Conv 5 & -0.200 $\pm$ 0.190 & 0.505 $\pm$ 0.186 \\
        
        FC & -0.019 $\pm$ 0.070 & 0.655 $\pm$ 0.156 \\
        
        \midrule
        All & -0.043 $\pm$ 0.212 & 0.401 $\pm$ 0.197 \\
        \bottomrule
    \end{tabular}
    \caption{Soft targets $\mathcal{\hat{B}}_Y$}
    \label{t:gradients_miniimagenet2}
  \end{subtable}
  \setlength{\belowcaptionskip}{-1em}
\caption{
MiniImageNet.
}
\label{t:gradients_miniimagenet}
\end{table}

\begin{table}[h]
  \centering
  \fontsize{6}{6}\selectfont
  \setlength{\tabcolsep}{3pt}
  \begin{subtable}[h]{0.45\textwidth}
  \centering
    \begin{tabular}{l c c}
        \toprule
        Layer & Corr. class 1 & Corr. class 2 \\
        \midrule

        FC 1 & 0.022 $\pm$ 0.178 & 0.255 $\pm$ 0.176 \\
        FC 2 & 0.012 $\pm$ 0.168 & 0.325 $\pm$ 0.228 \\
        FC 3 & 0.002 $\pm$ 0.264 & 0.675 $\pm$ 0.160 \\
        
        \midrule
        All & 0.012 $\pm$ 0.181 & 0.418 $\pm$ 0.181 \\
        \bottomrule
    \end{tabular}
    \caption{Hard targets (one-hot) $\mathcal{\hat{B}}_Y$}
    \label{t:gradients_mnist1}
  \end{subtable}
  \begin{subtable}[h]{0.45\textwidth}
    \centering
    \begin{tabular}{l c c}
        \toprule
        Layer & Corr. class 1 & Corr. class 2 \\
        \midrule
        FC 1 & 0.013 $\pm$ 0.182 & 0.055 $\pm$ 0.196 \\
        FC 2 & 0.015 $\pm$ 0.207 & 0.072 $\pm$ 0.246 \\
        FC 3 & 0.004 $\pm$ 0.311 & 0.100 $\pm$ 0.293 \\
        
        \midrule
        All & 0.011 $\pm$ 0.216 & 0.076 $\pm$ 0.233 \\
        \bottomrule
    \end{tabular}
    \caption{Soft targets $\mathcal{\hat{B}}_Y$}
    \label{t:gradients_mnist2}
  \end{subtable}
 \setlength{\belowcaptionskip}{-1em}
\caption{
MNIST.
}
\label{t:gradients_mnist}
\end{table}
\end{document}